\documentclass[11pt]{article}
\usepackage[dvips]{graphicx}
\usepackage{textcomp}
\usepackage{amsbsy}
\usepackage{latexsym}
\usepackage[mathscr]{eucal}
\usepackage{amsfonts,amsmath,amsthm}
\usepackage{amssymb}
\usepackage[usenames]{xcolor}
\usepackage{fullpage}

\begin{document}
\bibliographystyle{plain}
\title
{Neural networks: deep, shallow, or in between?}
\author{Guergana Petrova and Przemys{\l}aw Wojtaszczyk}
%
%
\maketitle

%
%
\newtheorem{lemma}{Lemma}[section]
\newtheorem{prop}[lemma]{Proposition}
\newtheorem{cor}[lemma]{Corollary}
\newtheorem{theorem}[lemma]{Theorem}
\newtheorem{remark}[lemma]{Remark}
\newtheorem{example}[lemma]{Example}
\newtheorem{definition}[lemma]{Definition}
\newtheorem{proper}[lemma]{Properties}
\newtheorem{assumption}[lemma]{Assumption}
%
\
\newcommand{\dI}{\Delta}
\newcommand\aconv{\mathop{\rm absconv}}


\abstract{We give estimates from below for the error of approximation of a compact subset from a Banach space by the outputs of feed-forward neural networks with width $W$, depth $\ell$ and 
Lipschitz activation functions. We show that, modulo logarithmic factors,  rates better that entropy numbers' rates are  possibly  attainable only for neural networks for which the depth $\ell\to\infty$,
 and that there is no gain 
 if we fix the depth and let the width $W\to\infty$.}

\section{Introduction}
\label{sec:1}
The fascinating new developments in the area of Artificial Intelligence (AI) and other important applications of neural networks prompt the  need for a theoretical mathematical study of their potential to reliably approximate complicated objects. Various network architectures have been  used in  different applications  with substantial  success rates without significant theoretical backing of the choices made. Thus, a natural question to ask is whether and how the architecture  chosen affects the approximation power of the outputs of the resulting neural network.
 
 In this paper, we  attempt to clarify how the width and the depth of a feed-forward neural network affect its worst performance. More precisely, we provide estimates from below for the error of approximation of a  compact subset $\mathcal K\subset X$ of a Banach space $X$ by the outputs of  feed-forward neural networks (NNs) with  width $W$, depth $\ell$, bound $w(W,\ell)$ on their parameters, and 
 Lipschitz activation functions. 
Note that  the ReLU function is included in our investigation since it is a Lipschitz function with a Lipschitz constant $L=1$.
 
 To prove our results, we assume that we know lower bounds on the  entropy numbers of the compact  sets $\mathcal K$ that we approximate by the outputs of feed-forward NNs.
Such  bounds are known for a wide range of classical and novel classes ${\mathcal K}$ and Banach spaces $X$, and are
usually of  the form $n^{-\alpha}[\log n]^\beta$,  $\alpha>0$, $\beta\in \mathbb R$. We refer the reader to 
 \cite[Chapters 3,4]{ET}, \cite[Chapter 15]{LGM},\cite [Section 5]{CDK}, \cite[Theorem  9]{JS}, or  \cite{CK,G}, where such examples are provided.
 
  It is a well known fact that the number $n$ of parameters of a feed-forward NN with width $W$ and depth $\ell$  is
 \begin{equation}
 \label{parameters}
 n\asymp \begin{cases}
 W^2\ell, \quad \hbox{when}\quad \ell>1,\\
 W, \quad\quad \hbox{when}\quad\ell=1.
 \end{cases}
 \end{equation}
 Let us  denote by $\Sigma(W,\ell,\sigma;w)$ the set of functions that are outputs of a such a NN with bounds $w=w(W,\ell)$ on its parameters and with Lipschitz activation function. 
 We prove estimates from below for  the error 
$
E(\mathcal K,\Sigma(W,\ell,\sigma;w))_X 
$
of approximation of a class $\mathcal K$ by the functions from $\Sigma(W,\ell,\sigma;w)$, see Theorem \ref{mainc}.
Our conclusion is that  under a moderate growth of the bound $w\asymp n^{\delta}$, $\delta\geq 0$, one  can possibly obtain rates of  approximation that are better 
than the corresponding entropy numbers' rates only when the 
depth of the NN is let to grow. 
If the rate of approximation of $\mathcal K$ by outputs of  feed-forward  NNs is  better than the decay rate of  its entropy numbers, then we say that we have super 
convergence. In fact, since we only obtain estimates from below, we claim that super  convergence is possibly attainable in such cases.
If the depth $\ell$ is fixed, then the rates of decay of $E(\mathcal K,\Sigma(W,\ell,\sigma;w))_X$ cannot be better (modulo logarithmic factors) than the rates
of the entropy numbers of $\mathcal K$. If both the width $W$ and depth $\ell$ are allowed to grow, then an improvement of the rates of decay of $E(\mathcal K,\Sigma(W,\ell,\sigma;w))_X$
 in comparison to the entropy numbers' decay is possible. Of course, the bound $w$ on the NN's parameters also has an effect and a fast growing bound, for example $w\asymp 2^n$,  could lead to 
 improved convergence in all cases. However,  one needs to be aware of the fact that  NNs with such bounds are  computationally infeasible.

We show  that the mapping assigning to each choice of neural network parameters the function that is an output of a feed-forward 
NN with these parameters is a Lipschitz mapping, see Theorem \ref{NN}. This allows us to study the  approximation properties of such NNs via the recently introduced 
Lipschitz widths, see \cite{PW,JMLR}. We have utilized this approach in \cite{JMLR} to discuss deep ($W=W_0$ is fixed and  $\ell\to\infty$) 
and shallow ($W\to\infty$ and  $\ell=1$) NNs with bounded Lipschitz or ReLU activation functions and their limitations in approximating compact 
sets $\mathcal K$. Here, we implement the developed technique to treat NNs for which both  $W,\ell\to \infty$. Results in this direction are available 
for shallow and deep NNs, and we refer the reader to the series 
of works \cite{YB,BHLM,YZ,Y1,Y2,SYZ,AFGM,DHP,M,MMR}, where various estimates from below are given for the error of approximation  for particular  classes 
 $\mathcal K$ and Banach spaces $X$.

The paper is organized as follows. In \S \ref{S2},  we introduce our notation, recall the definitions of NNs, entropy numbers and Lipschitz 
widths, and state some known results about them.
 We show in \S\ref{S3} that feed-forward NNs  are Lipschitz mappings. Finally, in \S\ref{S4}, we use results for Lipschitz widths to derive 
 estimates from below for the error of neural network approximation for a compact class $\mathcal K$.

\section{Preliminaries}
\label{S2}
In this section, we introduce our notation and recall some known facts about NNs, Lipschitz widths and entropy numbers. In what follows, we will denote by  
 $A\gtrsim B$ the fact that there is an absolute constant $c>0$ such that $A\geq cB$, where $A,B$ are some expressions that depend on some variable which tends to infinity.
 Note that  the value of $c$ may change from line to line, but is always independent on that variable. Similarly, we 
use the notation $A\lesssim B$ (defined in an analogues way)  and $A\asymp B$ if $A\gtrsim B$ and $A\lesssim B$. 

 We also write  $A=A(B)$ to stress  the fact that  the quantity $A$ depends on  $B$. For example, if $C$ is a constant, the expression 
 $C=C(d,\sigma)$ means that $C$ depends on 
 $d$ and $\sigma$.

\subsection{Entropy numbers}
\label{subsec:2}
We recall, see  e.g. \cite{C, CS, LGM}, that
the {\it entropy numbers} $\epsilon_n({\mathcal K}) _X$, $n\geq 0$, of a compact set ${\mathcal K}\subset X$ are defined 
as the infimum of all $\epsilon>0$ for which $2^n$ balls with centers from $X$ and radius $\epsilon$ cover ${\mathcal K}$.  
Formally,  we  write
$$ 
\epsilon_n({\mathcal K})_X=\inf\{ \epsilon>0 \ :\ {\mathcal K} 
\subset \bigcup_{j=1}^{2^n} B(g_j,\epsilon), \ g_j\in X, \ j=1,\ldots,2^n\}.
$$

\subsection{Lipschitz widths}
We denote by    $({\mathbb R}^n,\|.\|_{Y_n})$, $n\in \mathbb N$,
the $n$-dimensional Banach space with a fixed norm $\|\cdot\|_{Y_n}$, by
$$
 B_{Y_n}(r):=\{y\in {\mathbb R}^n:\,\,\|y\|_{Y_n}\leq r\},
 $$
its  ball with radius $r$, and by
$$
\|y\|_{\ell_\infty^n}:=\max_{j=1,\ldots,n}|y_j|, 
$$
the $\ell_\infty$ norm of  $y=(y_1,\ldots,y_n)\in \mathbb R^n$.
 The  Lipschitz widths $d^\gamma_n({\mathcal K})_X$ of the compact set $\mathcal K$ with respect to the norm $\|\cdot\|_X$ is  defined  as
\begin{equation}
\label{pp1}
d^\gamma_n({\mathcal K})_X:=\inf_{{\mathcal L}_n, \,r>0,\,\|\cdot\|_{Y_n}}\,\,\,\sup_{f\in {\mathcal K}}\,\,\inf_{y\in B_{Y_n}(r)} \|f-{\mathcal L}_n(y)\|_X,
\end{equation}
where the  infimum is taken over all $\gamma/r$-Lipschitz maps 
${\mathcal L}_n:(B_{Y_n}(r),\|\cdot\|_{Y_n})\to X$,  all $r>0$, and all norms $\|\cdot\|_{Y_n}$ in $\mathbb R^n$. We have proven, see Theorem 9 in 
\cite{JMLR}, the following result which relates the behavior of the entropy numbers of $\mathcal K$ and its Lipschitz widths with a Lipschitz constant $\gamma=2^{\varphi(n)}$.

\begin{theorem}
\label{widthsfrombelownew}  
For any  compact set ${\mathcal K}\subset X$, we consider the Lipschitz width $d_n^{\gamma_n}({\mathcal K})_X$ with 
 Lipschitz constant  $\gamma_n=2^{\varphi(n)}$, where   $\varphi(n)\geq c\log_2n$ for some fixed constant $c>0$. Let $\alpha>0$ and $\beta\in\mathbb R$. Then the following holds:
\begin{equation} 
\label{widths(i)gen}
{\rm (i)}\,\,\epsilon_n({\mathcal K})_X\gtrsim \frac{(\log_2 n)^\beta}{n^{\alpha}},\quad n\in \mathbb N\quad  \Rightarrow\quad d_n^{\gamma_n}({\mathcal K})_X\gtrsim
\frac{[\log_2 (n\varphi(n))]^{\beta}}{[n\varphi(n)]^{\alpha}}, \quad n\in \mathbb N;
\end{equation}
\begin{equation} 
\label{widths(ii)gen}
{\rm (ii)}\,\,\epsilon_n({\mathcal K}) _X\gtrsim [\log_2 n]^{-\alpha},\quad n\in \mathbb N\Rightarrow\quad d_n^{\gamma_n}({\mathcal K})_X\gtrsim [\log_2 (n\varphi(n))]^{-\alpha}, \quad n\in \mathbb N.
\end{equation} 
\end{theorem}

\subsection{Neural networks}
Let us  denote by $C(\Omega)$ the set of continuous functions defined on the compact set $\Omega\subset \mathbb R^d$, equipped with the uniform norm.

A feed-forward NN with activation function 
$\sigma:\mathbb R\to\mathbb R$, width $W$, depth $\ell$ and bound $w=w(W,\ell)$ on its parameters  generates a family $\Sigma(W,\ell,\sigma;w)$ of continuous functions  
$$ 
\Sigma(W,\ell,\sigma;w):=\{\Phi^{W,\ell}_\sigma(y): \,\,y\in \mathbb R^{ n}\} \subset C(\Omega), \quad \Omega\subset \mathbb R^d,
$$ 
where the number of parameters $n$ satisfies (\ref{parameters}).
Each $y\in \mathbb R^{n}$, $\|y\|_{\ell^n_\infty}\leq w$  determines a continuous function $\Phi^{W,\ell}_\sigma(y)\in \Sigma(W,\ell,\sigma;w)$, defined on $\Omega$,  of the form
\begin{equation}
\label{NN1}
\Phi^{W,\ell}_\sigma(y):=A^{(\ell)}\circ\bar\sigma\circ A^{(\ell-1)}\circ\ldots\circ \bar\sigma\circ A^{(0)},
\end{equation}
where  $\bar\sigma:\mathbb R^W\to\mathbb R^W$ is given  by 
\begin{equation}
\label{bars}
\bar\sigma(z_{1},\ldots,z_{W})=(\sigma(z_{1}),\ldots,\sigma(z_{W})),
\end{equation}
and  $A^{(0)}:\mathbb R^d\to\mathbb R^W$, $A^{(j)}:\mathbb R^{W}\to\mathbb R^{W}$, $j=1,\ldots,\ell-1$,  and $A^{(\ell)}:\mathbb R^W\to\mathbb R$ are affine mappings.
The coordinates of  $y\in\mathbb R^{n}$  are  the entries of the matrices and offset   vectors (biases) of the affine mappings $A^{(j)}$,
$j=0,\ldots,\ell$, taken in a pre-assigned order. The entries of $A^{(j)}$
 appear before those of $A^{(j+1)}$ and the ordering  for each $A^{(j)}$ is done in  the same way. 
 We refer the reader to \cite{DHP} and the references therein for detailed study of such NNs with fixed width $W=W_0$ and depth $\ell\to\infty$.

We view a feed-forward NN as a mapping that to each vector of parameters  $y\in \mathbb R^{n}$ 
assigns the output $\Phi^{W,\ell}_\sigma(y)\in \Sigma(W,\ell,\sigma;w)$ of this network,
\begin{equation}
\label{map}
y\to \Phi^{W,\ell}_\sigma(y), 
\end{equation}
where all parameters (entries of the matrices and biases) are bounded by $w(W,\ell)$,
namely
$$
\Sigma(W,\ell,\sigma;w)=\Phi^{W,\ell}_\sigma(B_{\ell_\infty^{n}}(w(W,\ell))),
$$
with $\Phi^{W,\ell}_\sigma$ being defined in (\ref{NN1}).

Lower bounds for the error of approximation of a class $\mathcal K\subset X$ by the outputs of  DNNs (when $W=W_0$ for a fixed $W_0$ and $\ell\to\infty$,  in which $n\asymp \ell$) and SNNs (when $\ell=1$ and $W\to \infty$,  in which $n\asymp W$) have been discussed in \cite{JMLR} in the case of bounded Lipschitz or ReLU activation functions. In this paper, we state similar results  for any feed-forward NN with 
general Lipschitz activation function. We use the approach from  \cite{JMLR}  and first  show that the mapping (\ref{map}) is a Lipschitz mapping.

\section{Feed-forward NNs are Lipshitz mappings}
\label{S3} 
Let us denote by 
\begin{equation}
\label{fas}
L:=\max\{L',|\sigma(0)|\},
\end{equation}
where $L'$ is the   Lipschitz constant of $\sigma$.
 Then
the following theorem is a generalization of Theorems 3 and 5 from \cite{JMLR} to the case of any feed-forward NN.
\begin{theorem}
  \label{NN}
 Let X be a Banach space such that $C([0,1]^d)\subset X$ is continuously embedded in $X$. Then the mapping $\Phi^{W,\ell}_\sigma:(B_{\ell_\infty^{n}}(w(W,\ell)), \|\cdot\|_{\ell_\infty^{n}})\to  X$, defined in 
  {\rm (\ref{NN1})} with a Lipschitz function $\sigma$,  is an $L_n$-Lipschitz mapping, that is, 
  $$
  \|\Phi^{W,\ell}_\sigma(y)-\Phi^{W,\ell}_\sigma(y')\|_X\leq  L_n\|y-y'\|_{\ell_\infty^{n}}, \quad y,y'\in B_{\ell_\infty^{n}}(w(W,\ell)).
    $$ 
Moreover, there are constants $c_1,c_2>0$ such that     
  $$
2^{c_1\ell\log_2(W(w+1)))}<L_n<2^{c_2\ell\log_2(W(w+1)))}, \quad w=w(W,\ell),
  $$
 provided  $LW\geq 2$.
 \end{theorem}
{\bf Proof:} 
Let us first set up the notation
 $
 \displaystyle{ \|g\|:=\max_{1\le i\le W} \|g_i\|_{C(\Omega)}},
 $
where $g$ is the vector function  $g=(g_1,\dots,g_W)^T$ whose coordinates $g_i\in C(\Omega)$. We also will use
$$
w:=w(W,\ell), \quad \hbox{and}\quad \tilde w:=w+1.
$$

  Let
    $y,y'$ be the two parameters from $B_{\ell_\infty^{n}}(w(W,\ell))$  that determine the 
    continuous functions $\Phi^{W,\ell}_\sigma(y), \,\Phi^{W,\ell}_\sigma(y')\in \Sigma(W,\ell,\sigma;w)$.  We fix $x\in \Omega$ and  denote  by 
$$
  \eta^{(0)}(x) := { \overline\sigma} (A_0x+b^{(0)}),\quad  \eta'^{(0)}(x) := { \overline\sigma} (A'_0x+b'^{(0)}),
  $$
$$  
\eta^{(j)}:= {\overline\sigma}(A_j\eta^{(j-1)}+b^{(j)}),\quad  \eta'^{(j)}:= {\overline\sigma}(A'_j{\eta}'^{(j-1)}+b'^{(j)}), \quad j=1,\ldots,\ell-1,
$$
$$
\eta^{(\ell)} := A_\ell\eta^{(\ell-1)}+b^{(\ell)},\quad  \eta'^{(\ell)}:=A'_\ell\eta'^{(\ell-1)}+b'^{(\ell)}.
$$
Note that  $A_0,A_0'\in \mathbb R^{W\times d}$, $A_j,A_j'\in \mathbb R^{W\times W}$, $b^{(j)},b'^{(j)}\in \mathbb R^W$, for $j=0,\ldots,\ell-1$,  while
$A_\ell,A_\ell'\in \mathbb R^{1\times W}$, and $b^{(\ell)},b'^{(\ell)}\in \mathbb R$.   
Each of the $\eta^{(j)}, \eta'^{(j)}$, $j=0, \ldots,\ell-1$, is a 
continuous vector function with $W$ coordinates, while  $\eta^{(\ell)}, \eta'^{(\ell)}$ are  the outputs of the NN with activation function $\sigma$ and parameters $y,y'$, respectively.  

Since, see (\ref{fas}), 
  $$
|\sigma(t)|\leq |\sigma(t)-\sigma(0)|+|\sigma(0)|\leq L(|t|+1), \quad |\sigma(t_1)-\sigma(t_2)|\leq L|t_1-t_2|, \quad t_1,t_2\in \mathbb R,
 $$
it follows that  for any $m$, vectors  $ \bar y,\hat y,\eta \in \mathbb R^m$ and  numbers $y_0,\hat y_0\in\mathbb R$, where $\bar y,y_0$ and $\hat y,\hat y_0$  are  subsets  of the coordinates  of 
 $y,y'\in \mathbb R^{n}$, respectively, we have
 \begin{eqnarray}
\label{si}
 |\sigma( \bar y\cdot \eta+y_0)|&\leq& L(|\bar y\cdot \eta+y_0|+1)\leq L(m\|\eta\|_{\ell_\infty^m}+1)\|y\|_{\ell_\infty^{n}}+L\\
 \nonumber
 &\leq&L(m\|\eta\|_{\ell_\infty^m}+1)w+L<L\tilde wm\|\eta\|_{\ell_\infty^m}+ L\tilde w
 \end{eqnarray}
 and
\begin{eqnarray}
  |\sigma(\bar y\cdot \eta+y_0)-\sigma(\hat y\cdot \eta+\hat y_0)|\leq L(m\|\eta\|_{\ell_\infty^m}+1)\|y-y'\|_{\ell_\infty^{n}}.
\end{eqnarray}
Then we have
$\|\eta'^{(0)}\|< L\tilde wd+L\tilde w$ (when $m=d$ and $\eta=x$) and 
$$
\|\eta'^{(j)}\|<LW\tilde w\|\eta'^{(j-1)}\|+L\tilde w, \quad j=1,\ldots,\ell,
$$
 (when  $m=W$ and $\eta=\eta'^{(j-1)}$).
 One can show by induction that for $j=1,\ldots,\ell$, 
$$
\|\eta'^{(j)}\|\leq dW^j[L\tilde  w]^{j+1}+L\tilde  w\sum_{i=0}^j[LW\tilde  w]^i.
$$
Therefore, we have that 
\begin{eqnarray}
\label{bound}
\|\eta'^{(j)}\|\leq dW^j[L\tilde  w]^{j+1}+2L\tilde  w[LW\tilde  w]^j=(d+2)L\tilde  w[LW\tilde  w]^j,
\end{eqnarray}
since $L W\tilde  w>LW\geq 2$.
The above inequality also holds for $j=0$.

 Clearly, we have
\begin{eqnarray}
\nonumber
 \|\eta^{(0)}-\eta^{'(0)}\|
 \leq 
  L(d+1)\|y-y'\|_{\ell_\infty^{n}}=:C_0\|y-y'\|_{\ell_\infty^{n}}.
\end{eqnarray}
     Suppose we have proved the inequality 
$$
 \|\eta^{(j-1)}-\eta'^{(j-1)}\|\leq C_{j-1}\|y-y'\|_{\ell_\infty^{ n}},
$$ 
for some constant $C_{j-1}$. 
Then we derive that
  \begin{eqnarray} 
  \nonumber
 \|\eta^{(j)}-\eta'^{(j)}\|&\leq& L \|A_j\eta^{(j-1)}+b^{(j)} -A_j'\eta'^{(j-1)}-b'^{(j)}\|
  \\ \nonumber
  &\le&   L\|A_j(\eta^{(j-1)}-\eta'^{(j-1)})\|+ L\|(A_j-A'_j)\eta'^{(j-1)}\|+L\|b^{(j)}-b'^{(j)}\|\\
  \nonumber
  &\leq &LW\|y\|_{\ell_\infty^{n}}\|\eta^{(j-1)}-\eta'^{(j-1)}\|
 +
LW\|y-y'\|_{\ell_\infty^{ n}}\|\eta'^{(j-1)}\|+L\|y-y'\|_{\ell_\infty^{ n}}\\
\nonumber
  &\le&
(LW\tilde wC_{j-1}+LW(d+2)L\tilde w[LW\tilde w]^{j-1}+L)\|y-y'\|_{\ell_\infty^{ n}}\\
\nonumber
  &=&L(W\tilde wC_{j-1}+(d+2)[LW\tilde w]^{j}+1)\|y-y'\|_{\ell_\infty^{ n}}\\
\nonumber
& =:&
  C_{j}\|y-y'\|_{\ell_\infty^{ n}},
  \end{eqnarray} 
  where  we  have used that $\|y\|_{\ell_\infty^{\tilde n}}\leq w$,  the bound (\ref{bound}), and the induction hypothesis. 
The relation between $C_j$ and $C_{j-1}$ can be written as
   $$
  C_0=L(d+1), \quad  C_{j}=L(W\tilde wC_{j-1}+(d+2)[LW\tilde w]^{j}+1), \quad j=1,\ldots,\ell.
$$
Clearly, 
$$
C_1=L((d+1)LW\tilde w+(d+2)LW\tilde w+1)< (d+2)L(2LW\tilde w+1),
$$
 and we obtain by induction that 
\begin{eqnarray}
\nonumber
  C_{\ell}&<&(d+2)L\left(\ell[LW\tilde w]^{\ell}+\sum_{i=0}^{\ell}[LW\tilde w]^{i}\right).
\end{eqnarray}
If we use the fact $2\leq  LW<LW\tilde w$, we derive the inequality
$$
 C_{\ell}
< (d+2)L(\ell+2)[LW\tilde w]^{\ell}.
$$
Finally, we have
\begin{eqnarray}
\nonumber
\|\Phi^{W,\ell}_\sigma(y)-\Phi^{W,\ell}_\sigma(y')\|_{C(\Omega)}&=&\|\eta^{(\ell)}-\eta'^{(\ell)}\| \leq C_{\ell}\|y-y'\|_{\ell_\infty^{n}}\\
\nonumber
&<&
(d+2)L(\ell+2)[LW\tilde w]^{\ell}\|y-y'\|_{\ell_\infty^{n}},
\end{eqnarray}
  and therefore
  $$
  \|\Phi^{W,\ell}_\sigma(y)-\Phi^{W,\ell}_\sigma(y')\|_X\leq c_0\|\Phi^{W,\ell}_\sigma(y)-\Phi^{W,\ell}_\sigma(y')\|_{C(\Omega)}\leq \tilde C\ell [LW\tilde w]^{\ell}\|y-y'\|_{\ell_\infty^{n}}, 
  $$
where $\tilde C=\tilde C(d,\sigma)$. Clearly,  the Lipschitz constant $L_n:=\tilde C\ell [LW\tilde w]^{\ell}$ is such that 
 $2^{c_1\ell\log_2(W(w+1))}<L_n<2^{c_2\ell\log_2(W(w+1))} $ for some $c_1,c_2>0$,
 and the proof is completed.
\hfill $\Box$
\bigskip

\begin{remark}
\label{Sremark}
Note that the proof of  Theorem {\rm \ref{NN}} holds also in the case when every coordinate  of $\bar \sigma$, see 
(\ref{bars}), is chosen to be a different Lipschitz function $\sigma$ as long as $LW\geq 2$, where $L$ is defined via (\ref{fas}).
 \end{remark}

\section{Estimates from below for neural network approximation}
\label{S4}
In this section,  we consider Banach spaces $X$ such that  $C([0,1]^d)$ is continuously embedded in $X$. 
Let us denote by 
$$
 E(f,\Sigma(W,\ell,\sigma;w))_X:=\inf_{y\in B_{\ell_\infty}^{n}(w)}
\|f-\Phi^{W,\ell}_\sigma(y)\|_X,
$$
the error of approximation in the norm $\|\cdot\|_X$ of the element $f\in \mathcal K$ by the set of outputs $\Sigma(W,\ell,\sigma;w)$ of a 
feed-forward NN with width $W$, depth $\ell$, activation function $\sigma$, and a bound $w$  on its parameters $y$, that is $\|y\|_{\ell_\infty^{n}}\leq w$.
We also denote by
$$
E(\mathcal K,\Sigma(W,\ell,\sigma;w))_X:=\sup_{f\in \mathcal K}\,\,E(f,\Sigma(W,\ell,\sigma;w))_X,
$$
the error for the class $\mathcal K\subset X$.
 It follows from  Theorem \ref{NN}  that 
\begin{equation}
\label{main}
E(\mathcal K,\Sigma(W,\ell,\sigma;w))_X\geq d^{\gamma_n}_n(\mathcal K)_X, \quad \hbox{with}\quad \gamma_n=2^{c\ell\log_2(W(w+1))}=:2^{\varphi(n)}, 
\end{equation}
for some $c>0$.
Therefore, see (\ref{parameters}), 
$$
n\varphi(n)=
\begin{cases}
cn\ell\log_2(W(w+1)), \quad n\asymp W^2\ell,\quad \ell>1,\\
cn\log_2(n(w+1)), \quad n\asymp W, \quad \ell=1,
\end{cases}
$$
and we can state the following corollary of  (\ref{main}) and Theorem \ref{widthsfrombelownew}.

\begin{theorem}
\label{mainc}
Let $\Sigma(W,\ell,\sigma;w)$ be the set of outputs of an $n$ parameter  NN with width $W$, 
depth  $\ell$, Lipschitz activation function $\sigma$ and  weights and biases bounded by $w$, where $LW\geq 2$.
Then, the  error of approximation  $E(\mathcal K,\Sigma(W,\ell,\sigma;w))_X$ of a compact subset
 ${\mathcal K}$ of a Banach space $X$ by $\Sigma(W,\ell,\sigma;w)$ satisfies the following 
  estimates from below,  provided we know the following information about the entropy numbers $\epsilon_n({\mathcal K})_X$ of $\mathcal K$:
\begin{itemize}
\item if for $\alpha>0$ and $\beta\in \mathbb R$ we have
\begin{equation} 
\nonumber
\epsilon_n({\mathcal K})_X\gtrsim  \frac{[\log_2 n]^\beta}{n^{\alpha}}, \, n\in \mathbb N,
\end{equation} 
then 
\begin{equation}
\nonumber
E(\mathcal K,\Sigma(W,\ell,\sigma;w))_X\gtrsim 
\begin{cases}
\frac{1}{n^\alpha\ell^\alpha}\cdot \frac{[\log_2 (n\ell\log_2(W(w+1)))]^{\beta}}{[\log_2(W(w+1))]^{\alpha}},\quad n\asymp W^2\ell,\quad \ell>1,\\\\
\frac{1}{n^\alpha}\cdot \frac{[\log_2 (n\log_2(nw))]^{\beta}}{[\log_2(n(w+1))]^{\alpha}}, \quad \quad\quad\quad\quad n\asymp W, \quad \ell=1.
\end{cases}
\end{equation}
\item if for $\alpha>0$ we have
\begin{equation} 
\nonumber
\epsilon_n({\mathcal K}) _X\gtrsim [\log_2 n]^{-\alpha}, \,n\in \mathbb N,
\end{equation}
then
\begin{equation}
\nonumber
E(\mathcal K,\Sigma(W,\ell,\sigma;w))_X\gtrsim 
\begin{cases}
[\log_2 (n\ell\log_2(W(w+1)))]^{-\alpha},\quad n\asymp W^2\ell,\quad \ell>1,\\\\
[\log_2 (n\log_2(n(w+1)))]^{-\alpha}, \quad\quad\quad n\asymp W, \quad \ell=1.
\end{cases}
\end{equation} 
\end{itemize}
\end{theorem}
{\bf Proof:}
 The proof follows directly from (\ref{main}) and Theorem \ref{widthsfrombelownew}.
\hfill $\Box$

\begin{remark}
Theorem \ref{mainc} gives various estimates from below depending on the behavior of the bound $w=w(W,\ell)$ on the absolute values of the parameters of the NN.
Here we state only one particular case. Under the conditions of Theorem \ref{mainc} with  $w=w(W,\ell)={\rm const}$, we have: 
\begin{itemize}
\item if for $\alpha>0$ and $\beta\in \mathbb R$ we have
\begin{equation} 
\nonumber
\epsilon_n({\mathcal K})_X\gtrsim  \frac{[\log_2 n]^\beta}{n^{\alpha}}, \, n\in \mathbb N,
\end{equation} 
then 
\begin{equation}
\nonumber
E(\mathcal K,\Sigma(W,\ell,\sigma;w))_X\gtrsim 
\begin{cases}
\frac{1}{n^\alpha\ell^\alpha}\cdot \frac{[\log_2 (n\ell\log_2W)]^{\beta}}{[\log_2W]^{\alpha}},\quad n\asymp W^2\ell,\quad \ell>1,\\\\
\frac{1}{n^\alpha}\cdot [\log_2 n]^{\beta-\alpha}, \quad\quad\quad\quad \,\,n\asymp W, \quad \ell=1.
\end{cases}
\end{equation}
\item if for $\alpha>0$ we have
\begin{equation} 
\nonumber
\epsilon_n({\mathcal K}) _X\gtrsim [\log_2 n]^{-\alpha}, \,n\in \mathbb N,
\end{equation}
then
\begin{equation}
\nonumber
E(\mathcal K,\Sigma(W,\ell,\sigma;w))_X\gtrsim 
\begin{cases}
[\log_2 (n\ell\log_2W)]^{-\alpha},\quad n\asymp W^2\ell,\quad \ell>1,\\\\
[\log_2 n]^{-\alpha},  \quad\quad\quad\quad\quad\quad n\asymp W, \quad \ell=1.
\end{cases}
\end{equation} 
\end{itemize}
\end{remark}

\bigskip 

{\bf Acknowledgments:} 
G.P. was supported by the NSF Grant DMS 2134077  and ONR Contract N00014-20-1-278.

\vskip .1in
\noindent
{\bf Affiliations:}

\noindent
Guergana Petrova, Department of Mathematics, Texas A$\&$M University, College Station, TX 77843,  gpetrova$@$math.tamu.edu.
\vskip .1in
\noindent
Przemys{\l}aw Wojtaszczyk, Institut of Mathematics,  Polish Academy of Sciences, ul. 
{\'S}niadeckich 8,  00-656 Warszawa, Poland, wojtaszczyk$@$impan.pl

\end{document}